\documentclass[runningheads]{llncs}
\usepackage[T1]{fontenc}
\usepackage{graphicx}
\usepackage[babel,german=quotes]{csquotes}
\usepackage{booktabs}
\usepackage{array}
\usepackage{amsmath}
\usepackage[table]{xcolor}
\usepackage{soul}
\usepackage{url}
\begin{document}
\title{Multimodal Late Fusion Model for Problem-Solving Strategy Classification in a Machine Learning Game}
\titlerunning{Multimodal Late Fusion Model for Problem-Solving Strategy Classification}

\author{Clemens Witt\inst{1}\orcidID{0009-0005-8160-4029} \and
Thiemo Leonhardt\inst{2}\orcidID{0000-0003-4725-9776} \and
Nadine Bergner\inst{2}\orcidID{0000-0003-3527-3204} \and
Mareen Grillenberger\inst{1}\orcidID{0000-0002-8477-1464}}

\authorrunning{C. Witt et al.}

\institute{TUD Dresden University of Technology \and RWTH Aachen University}

\maketitle

\begin{abstract}
Machine learning models are widely used to support stealth assessment in digital learning environments. 
Existing approaches typically rely on abstracted gameplay log data, which may overlook subtle behavioral cues linked to learners’ cognitive strategies. 
This paper proposes a multimodal late fusion model that integrates screencast-based visual data and structured in-game action sequences to classify students’ problem-solving strategies. 
In a pilot study with secondary school students $(N=149)$ playing a multitouch educational game, the fusion model outperformed unimodal baseline models, increasing classification accuracy by over 15\%. 
Results highlight the potential of multimodal ML for strategy-sensitive assessment and adaptive support in interactive learning contexts.
\end{abstract}

\section{Introduction}

Digital game-based learning (DGBL) environments offer distinct potential for embedding formative assessment within interactive learning processes.
By leveraging the immersive and data-rich nature of such environments, adaptive instruction can be enabled through continuous behavioral analysis – without disrupting learner engagement \cite{dicerboFutureAssessmentTechnologyRich2016}.
Recent advances in machine learning (ML) have opened up new possibilities for using gameplay data to infer learner competencies and strategies. However, many current approaches still rely heavily on expert-engineered features extracted from interaction logs. While effective in isolating predefined constructs, such abstractions risk overlooking subtle behavioral indicators; particularly those reflecting problem-solving behaviors and cognitive engagement.
To address this limitation, we present a multimodal ML approach for classifying students’ problem-solving strategies in a digital learning game designed to teach core concepts of decision tree learning.
The model integrates two complementary data modalities: (1) visual information extracted from gameplay screencasts, capturing non-verbal and temporal behavioral cues, and (2) symbolic representations of in-game actions. 
These modalities are combined using a late fusion architecture \cite{baltrusaitisMultimodalMachineLearning2019} to enhance both classification accuracy and interpretability.
Gameplay sequences were annotated according to a validated coding scheme distinguishing exploratory from structured problem-solving strategies, forming the basis for model training and evaluation.
The findings underscore the potential of multimodal models for strategy-sensitive assessment in digital learning environments and demonstrate the viability of screencast video as a primary data source for developing adaptive, behavior-aware support systems.
All data, model specifications, and training artifacts are publicly accessible via the Open Science Framework to ensure transparency and support replication.\footnote[1]{\url{https://osf.io/d8kec/}}

\section{Theoretical Background} \label{Theorectial_Background}

The interactive nature of DGBL environments allows formative assessment to be embedded directly into learning processes, enabling the continuous collection of learning-relevant data without disrupting immersion \cite{dicerboFutureAssessmentTechnologyRich2016}.
A prominent approach in this domain is \textit{Stealth Assessment}, which integrates assessment seamlessly into gameplay to infer learner competencies from behavioral data \cite{shuteStealthAssessmentMeasuring2013}.
Stealth assessment is grounded in the framework of Evidence-Centered Design (ECD), which systematically aligns learning tasks with evidence and competency models \cite{mislevyFocusArticleStructure2003}. ECD emphasizes the design of tasks that elicit observable behaviors, which in turn provide evidence for underlying competencies.

Recent research has increasingly leveraged ML to automate the extraction and modeling of assessment-relevant features, moving beyond expert-crafted models. This shift is driven by the growing availability of fine-grained interaction data and the limitations of manual feature engineering in capturing complex learning processes \cite{luScalableFlexibleInterpretable2024}.
While many studies have focused on predicting learning outcomes from gameplay data \cite{guptaMultimodalMultiTaskStealth2021,minDeepStealthGameBasedLearning2020}, more recent efforts seek to model cognitive processes and problem-solving strategies in greater depth \cite{akramImprovingStealthAssessment}.
To model the temporal dynamics of learning, ML architectures based on recurrent neural networks (RNN), such as Long Short-Term Memory (LSTM) networks \cite{hochreiterLongShorttermMemory1997a}, have shown strong performance in modeling sequential gameplay data \cite{choLearningPhraseRepresentations2014,rajendranStealthAssessmentStrategy2022}.
Additionally, multimodal learning analytics approaches integrate heterogeneous data sources such as clickstreams, textual inputs, and physiological signals to build more comprehensive and process-oriented learner models \cite{bliksteinMultimodalLearningAnalytics2016,hendersonSensorbasedDataFusion}.
\section{Methodology}

\subsection{Dataset}

The proposed classification model builds on a dataset obtained from a previous study $(N=92)$ that investigated students’ problem-solving strategies in the digital learning game \textit{Match the Monkeys}\footnote[2]{\url{https://ai.ddi.education/games/match-the-monkeys/}} \cite{wittGuessworkGamePlan2024a}.
In this game, learners construct decision trees by assembling puzzle-like filter elements to classify cartoon monkeys based on their visual features. 
To increase the representation of structured problem-solving strategies, 57 additional gameplay sessions were collected during introductory ML workshops with secondary school students (grades 8–10). 
In total, 149 gameplay sessions were recorded as screencasts, yielding approximately 30 hours of video data. 
Each session was annotated using a validated coding scheme developed in the previous study based on strategy typologies derived from programming education.
The scheme differentiates between exploratory strategies (\textit{Thoughtful Tinkering}, \textit{Trial \& Error Tinkering}) and structured approaches (\textit{Building}, \textit{Top-Down Structuring}, \textit{Bottom-Up Structuring}).
For model training, the annotated data were consolidated into three target classes: \textit{Thoughtful Tinkering}, \textit{Trial \& Error Tinkering}, and a combined class \textit{Structured Problem-Solving}. 
The resulting dataset consists of 380 samples from the \textit{Structured Problem-Solving} category, balanced by an equal number of randomly drawn instances from each of the two significantly more common exploratory categories.

\subsection{Model Architecture and Training}

\begin{figure}[t]
    \centering
    \includegraphics[width=\textwidth]{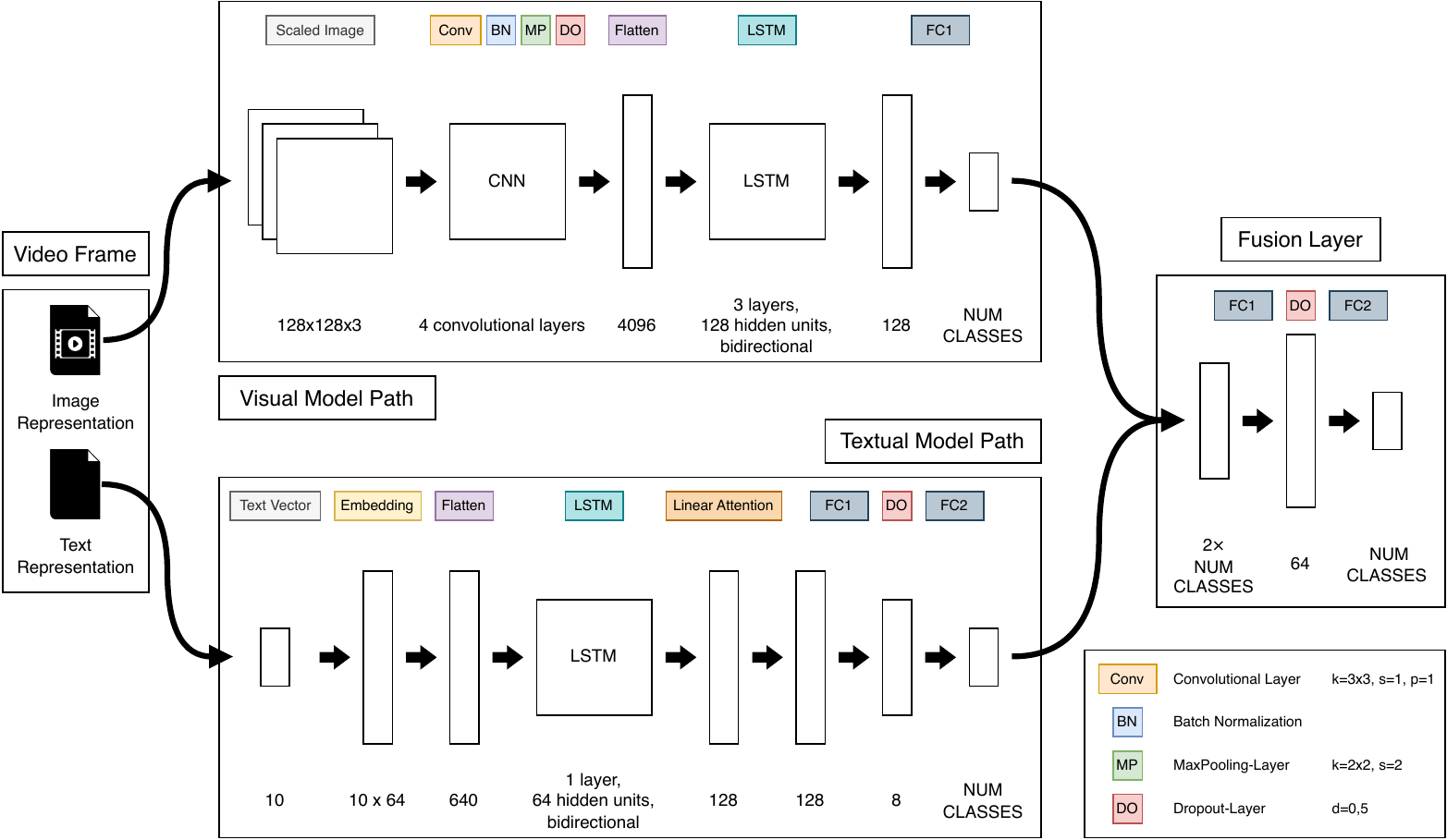}
    \caption{Late Fusion Multimodal Model Architecture}
    \label{fig:fusion-model}
\end{figure}

As illustrated in Figure \ref{fig:fusion-model}, the proposed classification model adopts a late fusion architecture \cite{baltrusaitisMultimodalMachineLearning2019} that combines two complementary input modalities: (1) visual data from gameplay screencasts capturing temporally embedded interaction patterns, and (2) symbolic representations of learners’ in-game decision trees. Each modality is processed independently via a dedicated subnetwork prior to decision-level integration.
The visual model path processes sequences of 60 screencast frames at a resolution of 128$\times$128 pixels. A four-layer convolutional neural network (CNN) extracts spatial features, which are passed to a bidirectional LSTM (BiLSTM) network to capture temporal dependencies. A final dense layer projects the output into the target class space.
In parallel, the textual model path processes symbolic sequences extracted via optical character recognition (OCR) from screencast frames. Filter texts are mapped to discrete token IDs, embedded into a continuous vector space, and fed into a BiLSTM network with a linear attention mechanism, highlighting semantically salient segments of the interaction sequences. The resulting representation is transformed through two fully connected layers to align with the classification target space.
The outputs of both unimodal pathways are concatenated in a fusion module comprising two dense layers. 

Model training followed a two-stage procedure. Both unimodal subnetworks were pretrained independently using the standard Adam optimizer along with dropout (visual: $d=0.3$, textual: $d=0.5$), L2 regularization (\texttt{weight\_decay} = $10^{-3}$ for visual, $10^{-2}$ for textual), batch normalization (visual only), and gradient clipping (\texttt{max\_norm} = 1.0).
Learning rate schedules were dynamically adjusted based on validation loss. The visual model used a batch size of 16, while the less complex textual model employed a batch size of 32.
Following pretraining, the outputs of the frozen submodels were used to train the fusion layer. 
This approach ensured stable, modality-specific representations and allowed the fusion layer to focus exclusively on optimizing cross-modal integration. 
\section{Results} \label{results}

To evaluate the performance of the proposed multimodal classification model, its accuracy and F1-scores were compared to the two unimodal baseline models. All models were trained repeatedly under identical conditions to control for performance variance due to random initialization and regularization – factors known to affect neural sequence models, particularly when trained on limited data \cite{goodfellowSequenceModelingRecurrent2016}.
Rather than specifying an arbitrary number of training runs, we applied a dynamic evaluation procedure based on confidence interval (CI) convergence. Assuming normally distributed accuracy values across independent training runs, the required number of replications to achieve a target CI width $B$ can be estimated as:
\begin{equation}
    N = (\frac{2 \sigma z}{B})^2 \hspace{0.3cm} \text{with } z = 1.96 \text{ for 95\% CI}
\end{equation}
We specified a margin of error $B=0.01 \; (\pm 0.5\%)$, triggering the evaluation pipeline to continue training until the CI narrowed to the target width. After 326 training runs, this criterion was met. A Shapiro-Wilk test $(p=0.125)$, along with a visual inspection using a histogram and Q–Q plot analysis, confirmed the normality of the accuracy distribution, validating the CI-based stopping condition.
The multimodal fusion model achieved a mean accuracy of 72.42\% (SD = 4.60\%, 95\% CI: $[0.7192, 0.7292]$), outperforming both the visual baseline (56.24\%, SD = 4.24\%) and the textual baseline (50.50\%, SD = 4.51\%). 
In terms of F1-score, the fusion model reached an overall value of 0.7538, demonstrating both class-wise robustness and generalizability. Particularly in the \textit{Structured Problem-Solving} class it achieved an F1-score of 0.88, compared to 0.69 for the visual model and 0.57 for the textual model. In both exploratory problem-solving strategy classes (\textit{Thoughtful Tinkering}, \textit{Trial \& Error Tinkering}), the fusion model also outperformed the baseline models across all submetrics.
Despite minimal feature engineering, the proposed model achieves performance levels comparable to state-of-the-art multimodal stealth assessment approaches \cite{guptaMultimodalMultiTaskStealth2021,hendersonSensorbasedDataFusion}. Its strong performance in differentiating structured from exploratory problem-solving strategies underscores its potential for adaptive, strategy-sensitive learner support in digital learning environments.

\begin{figure}[t]
    \centering
    \includegraphics[width=\linewidth]{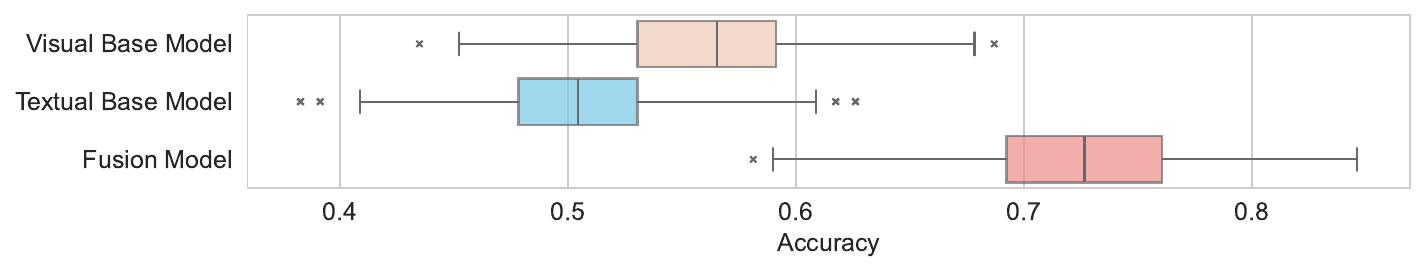}
    \caption{Accuracy Distribution of the Baseline Models and the Late Fusion Model}
    \label{fig:boxplot-accuracy}
\end{figure}

\section{Discussion}

This study proposed a multimodal late fusion model for classifying students’ problem-solving strategies in a digital learning game. By integrating visual gameplay screencasts with symbolic representations of in-game actions, the model substantially outperformed both unimodal baselines, particularly in identifying structured problem-solving strategies – an indicator of more effective learning behavior.
A central contribution of this work lies in employing screencast video as a primary data modality. In contrast to conventional gameplay log data, video captures rich, temporally embedded behavioral cues – such as element sequencing, pacing, and hesitation – that provide implicit insight into cognitive engagement. This approach advances stealth assessment by enabling strategy-level differentiation based on minimally preprocessed, generalizable data sources.
The reliable detection of exploratory vs. structured problem-solving strategies offers key opportunities for adaptive learning support. Learners exhibiting persistent trial-and-error behavior could be guided toward more systematic approaches through targeted prompts, while instructors could leverage real-time strategy profiles for diagnostic purposes. Adaptive feedback and game mechanics informed by strategy detection may further support productive learning trajectories.

Limitations of this study include the narrow domain context, reduced strategy set, and single-coder annotation  in the additionally collected sessions. Moreover, the model focuses solely on classification; motivational and causal dimensions remain to be adressed in future work.
Despite these constraints, the findings highlight the potential of multimodal modeling to enhance fine-grained, behavior-aware assessment in interactive learning environments and to inform the design of personalized, strategy-sensitive support systems.

\bibliographystyle{splncs04}

\begin{thebibliography}{10}
\providecommand{\url}[1]{\texttt{#1}}
\providecommand{\urlprefix}{URL }
\providecommand{\doi}[1]{https://doi.org/#1}

\bibitem{akramImprovingStealthAssessment}
Akram, B., Min, W., Wiebe, E., Mott, B., Boyer, K.E., Lester, J.: {Improving
  Stealth Assessment in Game-Based Learning with LSTM-based Analytics}  (2018)

\bibitem{baltrusaitisMultimodalMachineLearning2019}
Baltru{\v s}aitis, T., Ahuja, C., Morency, L.P.: Multimodal {Machine Learning}:
  {A Survey} and {Taxonomy}. IEEE Transactions on Pattern Analysis and Machine
  Intelligence  \textbf{41}(2),  423--443 (2019).
  \doi{10.1109/TPAMI.2018.2798607}

\bibitem{bliksteinMultimodalLearningAnalytics2016}
Blikstein, P., Worsley, M.: Multimodal {Learning Analytics} and {Education Data
  Mining}: Using computational technologies to measure complex learning tasks.
  Journal of Learning Analytics  \textbf{3}(2),  220--238 (2016).
  \doi{10.18608/jla.2016.32.11}

\bibitem{choLearningPhraseRepresentations2014}
Cho, K., van Merrienboer, B., Gulcehre, C., Bahdanau, D., Bougares, F.,
  Schwenk, H., Bengio, Y.: Learning {Phrase Representations} using {RNN
  Encoder-Decoder} for {Statistical Machine Translation} (2014).
  \doi{10.48550/arXiv.1406.1078}

\bibitem{dicerboFutureAssessmentTechnologyRich2016}
DiCerbo, K.E., Shute, V., Kim, Y.J.: The {Future} of {Assessment} in
  {Technology-Rich Environments}: {Psychometric Considerations}. In: Spector,
  M.J., Lockee, B.B., Childress, M.D. (eds.) Learning, {Design}, and
  {Technology}, pp. 1--21. Springer International Publishing, Cham (2016).
  \doi{10.1007/978-3-319-17727-4_66-1}

\bibitem{goodfellowSequenceModelingRecurrent2016}
Goodfellow, I., Bengio, Y., Courville, A.: Sequence modeling: Recurrent and
  recursive nets. In: Deep Learning, pp. 367--415. MIT press Cambridge, MA, USA
  (2016)

\bibitem{guptaMultimodalMultiTaskStealth2021}
Gupta, A., Carpenter, D., Min, W., Rowe, J., Azevedo, R., Lester, J.:
  Multimodal {Multi-Task Stealth Assessment} for {Reflection-Enriched
  Game-Based Learning}  (2021)

\bibitem{hendersonSensorbasedDataFusion}
Henderson, N.L., Rowe, J.P., Mott, B.W., Lester, J.C.: Sensor-based {Data
  Fusion} for {Multimodal Affect Detection} in {Game-based Learning
  Environments}

\bibitem{hochreiterLongShorttermMemory1997a}
Hochreiter, S., Schmidhuber, J.: Long short-term memory. Neural computation
  \textbf{9}(8),  1735--1780 (1997)

\bibitem{luScalableFlexibleInterpretable2024}
Lu, W., Laffey, J., Sadler, T.D., Griffin, J., Goggins, S.P.: A {Scalable},
  {Flexible}, and {Interpretable Analytic Pipeline} for {Stealth Assessment} in
  {Complex Digital Game-Based Learning Environments}: {Towards
  Generalizability}. Journal of Educational Data Mining  \textbf{16}(2),
  214--303 (Dec 2024). \doi{10.5281/zenodo.14503598}

\bibitem{minDeepStealthGameBasedLearning2020}
Min, W., Frankosky, M.H., Mott, B.W., Rowe, J.P., Smith, A., Wiebe, E., Boyer,
  K.E., Lester, J.C.: Deepstealth: Game-based learning stealth assessment with
  deep neural networks. IEEE Transactions on Learning Technologies
  \textbf{13}(2),  312--325 (2019)

\bibitem{mislevyFocusArticleStructure2003}
Mislevy, R.J., Steinberg, L.S., Almond, R.G.: Focus {Article}: {On} the
  {Structure} of {Educational Assessments}. Measurement: Interdisciplinary
  Research \& Perspective  \textbf{1}(1),  3--62 (Jan 2003).
  \doi{10.1207/S15366359MEA0101_02}

\bibitem{rajendranStealthAssessmentStrategy2022}
Rajendran, D., Prasanna, S.: {Stealth Assessment Strategy in Distributed
  Systems Using Optimal Deep Learning with Game Based Learning}. The Journal of
  Supercomputing  \textbf{78}(6),  8285--8301 (Apr 2022).
  \doi{10.1007/s11227-021-04236-y}

\bibitem{shuteStealthAssessmentMeasuring2013}
Shute, V., Ventura, M.: Stealth {Assessment}: {Measuring} and {Supporting
  Learning} in {Video Games}. The MIT Press (2013).
  \doi{10.7551/mitpress/9589.001.0001}

\bibitem{wittGuessworkGamePlan2024a}
Witt, C., Leonhardt, T., Marx, E., Bergner, N.: From {Guesswork} to {Game
  Plan}: {Exploring Problem-Solving-Strategies} in a {Machine Learning Game}.
  In: Pluh{\'a}r, Z., Ga{\'a}l, B. (eds.) Informatics in {Schools}. {Innovative
  Approaches} to {Computer Science Teaching} and {Learning}, vol. 15228, pp.
  73--84. Springer Nature Switzerland, Cham (2024).
  \doi{10.1007/978-3-031-73474-8_6}

\end{thebibliography}

\end{document}